\documentclass{article}

\usepackage{PRIMEarxiv}
\usepackage{cite}
\usepackage[utf8]{inputenc} 
\usepackage[T1]{fontenc}    
\usepackage{hyperref}       
\usepackage{url}            
\usepackage{booktabs}       
\usepackage{amsfonts}       
\usepackage{nicefrac}       
\usepackage{microtype}      
\usepackage{lipsum}
\usepackage{tikz}
\usepackage{fancyvrb}
\usepackage{booktabs}
\usetikzlibrary{positioning, shapes.geometric, arrows.meta}
\usepackage{fancyhdr}       
\usepackage{graphicx}       
\graphicspath{{media/}}     

\title{SpikeAtConv: An Integrated Spiking-Convolutional Attention Architecture for Energy-Efficient Neuromorphic Vision Processing}

\author{
  Wangdan Liao\\
  Beihang University  \\
  \texttt{liaowangdan@buaa.edu.cn} \\
  \And
  Weidong Wang\\
   Chinese PLA General Hospital\\
  \texttt{wangwd301@126.com} \\
}

\begin{document}
\maketitle

\begin{abstract}
Spiking Neural Networks (SNNs) offer a biologically inspired alternative to conventional artificial neural networks, with potential advantages in power efficiency due to their event-driven computation. Despite their promise, SNNs have yet to achieve competitive performance on complex visual tasks, such as image classification. This study introduces a novel SNN architecture designed to enhance computational efficacy and task accuracy. The architecture features optimized pulse modules that facilitate the processing of spatio-temporal patterns in visual data, aiming to reconcile the computational demands of high-level vision tasks with the energy-efficient processing of SNNs. 
Our evaluations on standard image classification benchmarks indicate that the proposed architecture narrows the performance gap with traditional neural networks, providing insights into the design of more efficient and capable neuromorphic computing systems.
\end{abstract}

\section{Introduction}

Spiking Neural Networks (SNNs) represent the forefront of a paradigm shift towards more energy-efficient and biologically plausible computational models. As the third generation of neural network technologies, SNNs offer a promising alternative to traditional machine intelligence systems by emulating the event-driven characteristics of biological neural processing \cite{maass1997networks}. The appeal of SNNs is multifaceted, with their ability not only to operate at lower power consumption, but also to perform computations in a manner that closely mirrors the spatiotemporal dynamics of the brain \cite{roy2019towards}. The spike-based communication protocol of SNNs is especially well-suited for sparse and asynchronous computations, making it highly appropriate for deployment on neuromorphic chips. These chips are designed to emulate the neural architecture of the brain, leveraging the inherent sparse activation patterns of SNNs to achieve significant energy efficiency improvements \cite{li2024brain, frenkel2023bottom, merolla2014million, davies2018loihi, pei2019towards}.

Despite their potential, SNNs have historically grappled with performance limitations, particularly in complex cognitive tasks that are easily handled by their ANN counterparts. This has prompted researchers to explore the adaptation of successful ANN architectures into the spiking domain. For instance, SNNs based on Convolutional Neural Networks (CNNs) have been developed, enabling the transposition of classic architectures like VGG and ResNet into SNN frameworks \cite{lecun1989backpropagation,wu2021progressive}. These adaptations have made significant strides, yet the quest for architectures that can fully exploit the unique advantages of SNNs continues.

The emergence of the Transformer architecture, originally designed for natural language processing, has sparked a new wave of innovations across various fields of machine learning \cite{vaswani2017attention}. Its success in ANNs has not gone unnoticed in the SNNs community, leading to the exploration of Transformer-based designs within spiking networks \cite{zhang2022spiking, zhou2023spikformer}. However, the integration of the self-attention mechanism into SNNs has been challenging, as it relies on operations that are at odds with the principles of spike-based processing, such as the energy-intensive Multiply-and-Accumulate (MAC) operations. Recent efforts have sought to reconcile this discrepancy by proposing spike-driven variants of the self-attention mechanism, aiming to retain the computational efficiency and low power consumption that are hallmarks of SNNs \cite{yao2023spike}. These innovations represent a significant departure from traditional Transformer models, yet the challenge remains to demonstrate their superiority over existing SNN designs in both performance and energy efficiency.

In this paper, we introduce an innovative spiking neural network framework called SpikeAtConv, designed to incorporate the strengths of advanced Transformer models into SNNs. An overview of the SpikeAtConv network is shown in Figure 1. Inspired by MaxViT, we propose a novel spike-driven transformer module named Spike-Driven Grid Attention. This module facilitates global spatial interactions within a single block, providing enhanced flexibility and efficiency compared to traditional spike-driven full self-attention or (shifted) window/local attention mechanisms. The SpikeAtConv Block, composed of Spike-Driven Grid Attention and ConvNeXt, serves as the core component of the SpikeAtConv network. Additionally, we have designed various Spiking Neuron (SPK) Blocks to enable a more flexible neuron activation mechanism, such as the Multi-Branch Parallel LIF SPK (MBPL) Block, which consists of multiple parallel neurons with different thresholds. The main contributions of this paper are as follows:

1. We design a series of SPK Blocks to explore the effects of multiple neurons with different thresholds and combinations on network performance. Through extensive experiments, we identify the optimal configuration of the spiking neuron module, which significantly enhances the computational performance of the model.

2. We develop Spike-Driven Grid Attention, enabling global spatial interactions within a single block. This allows the SpikeAtConv block to capture both local and global significant features  more effectively.

3. We propose the SpikeAtConv network, which is based on the developed SPK Block and SpikeAtConv Block. This architecture successfully adapts advanced transformer models to the SNN framework, thereby enhancing the computational performance and efficiency of the model.

4. Extensive eperiments show that the proposed model outperforms or is comparable to the state-of-the-art (SOTA) SNNs on the datasets. Notably, we achieved a top-1 accuracy of 81.23\% on ImageNet-1K using the directly trained Large SpikeAtConv , which is a SOTA result in the field of SNN.

\section{Related Work}\label{sec:related-work}

Our exploration of Spiking Neural Networks (SNNs) intersects with several key research domains, including the direct training of SNNs using surrogate gradients, as well as the integration of SNNs with advanced visual models such as Vision Transformers (ViTs).

ANN to SNN Conversion and Direct Training. The non-differentiable nature of spike functions has traditionally posed a challenge to applying conventional backpropagation to SNNs \cite{rumelhart1986learning}. In response to this, researchers have developed ANN-to-SNN conversion techniques that establish neuron equivalency, discretizing a trained ANN into an SNN \cite{deng2021optimal,hu2024advancing}. Simultaneously, direct training methods using surrogate gradients have been proposed, allowing for the training of SNNs despite the non-differentiability of the spiking process \cite{wu2018spatio,neftci2019surrogate}. Our work builds upon these foundations, utilizing direct training approaches due to their fine-grained temporal resolution and architecture-adaptive flexibility.

Advancements in Conv-based SNNs. The design of Conv-based SNNs has been significantly influenced by the concept of residual learning from ResNet \cite{he2016deep}. Efforts to extend SNNs to deeper architectures have led to the development of techniques such as tdBN \cite{zheng2021going}, as well as enhancements like SEW-Res-SNN and MS-Res-SNN \cite{fang2021deep,hu2024advancing}, which have facilitated the creation of SNNs with over 100 layers. 

Vision Transformers in SNNs. The impressive performance of ViTs \cite{dosovitskiy2021an} has prompted a shift from traditional convolutional neural networks to transformer-based architectures. Our work is particularly inspired by the Meta SpikeFormer architecture, which aims to bridge the gap between SNNs and ANNs in image classification tasks. While there have been advancements in ViTs such as PVT \cite{wang2021pyramid} and MLP Mixer \cite{tolstikhin2021mlp}, and explorations of the self-attention mechanism \cite{liu2021swin,chu2021twins}, their application within the SNN domain is still in its infancy. Our goal is to extend the capabilities of SNNs by incorporating self-attention mechanisms tailored to the spiking nature of these networks.

\section{Method}

\subsection{Overall Architecture}

In this study, we propose a spiking neural network architecture called SpikeAtConv, which is inspired by MaxViT  \cite{tan2019efficientnet,woo2023convnext,tu2022maxvit,dai2021coatnet}. The overall structure of the SpikeAtConv network is shown in Figure 1. MaxViT is a vision neural network architecture that effectively combines the strengths of Transformers and convolutional neural networks by integrating self-attention mechanisms with convolutional operations. Building on MaxViT, we modified both the Transformer and convolutional components to handle and generate spike signals, resulting in a novel SNN model.

Firstly, the Feature Extraction Layer of the model consists of two Convolutional Neural Network (CNN) layers and a spiking neuron (SPK) Block. Further details regarding the SPK Block will be provided in the following sections. The primary function of this layer is to downsample the input image, halving its resolution with a stride setting of 2 in the first CNN layer. Additionally, it converts continuous image data into neural spike signals, i.e., binary discrete data, making it suitable for subsequent processing.

Next is the Feature Encoding Layer of the model, which forms the core of the model. It includes four stages, each performing downsampling at the entrance to halve the resolution of the feature map, with no further downsampling within the same stage. Each stage consists of a series of SpikeAtConv Blocks, varying in number but collectively achieving deep feature encoding. The SpikeAtConv Blocks represent our novel integration of CNN, attention mechanisms, and SNN, designed to enhance the  performance of model. Detailed information about these modules is provided in subsequent sections. The depth of these four stages follows a spindle-shaped distribution; for instance, in the base model, the depths of the stages are 2, 6, 12, and 2, respectively. This design follows empirical rules of classification visual neural networks to effectively capture features and facilitate information flow.

Finally, the Decision Layer of the model is responsible for the classification task. It processes the output feature maps from the previous stages through global pooling, followed by a linear layer to predict the categories. This layer is designed to be both simple and efficient, capable of transforming complex, high-dimensional features into the final classification decision. The overall structure of our model leverages traditional methods while incorporating innovative SNN elements for enhanced performance.

\begin{figure}
  \centering
  \includegraphics[width=1\textwidth]{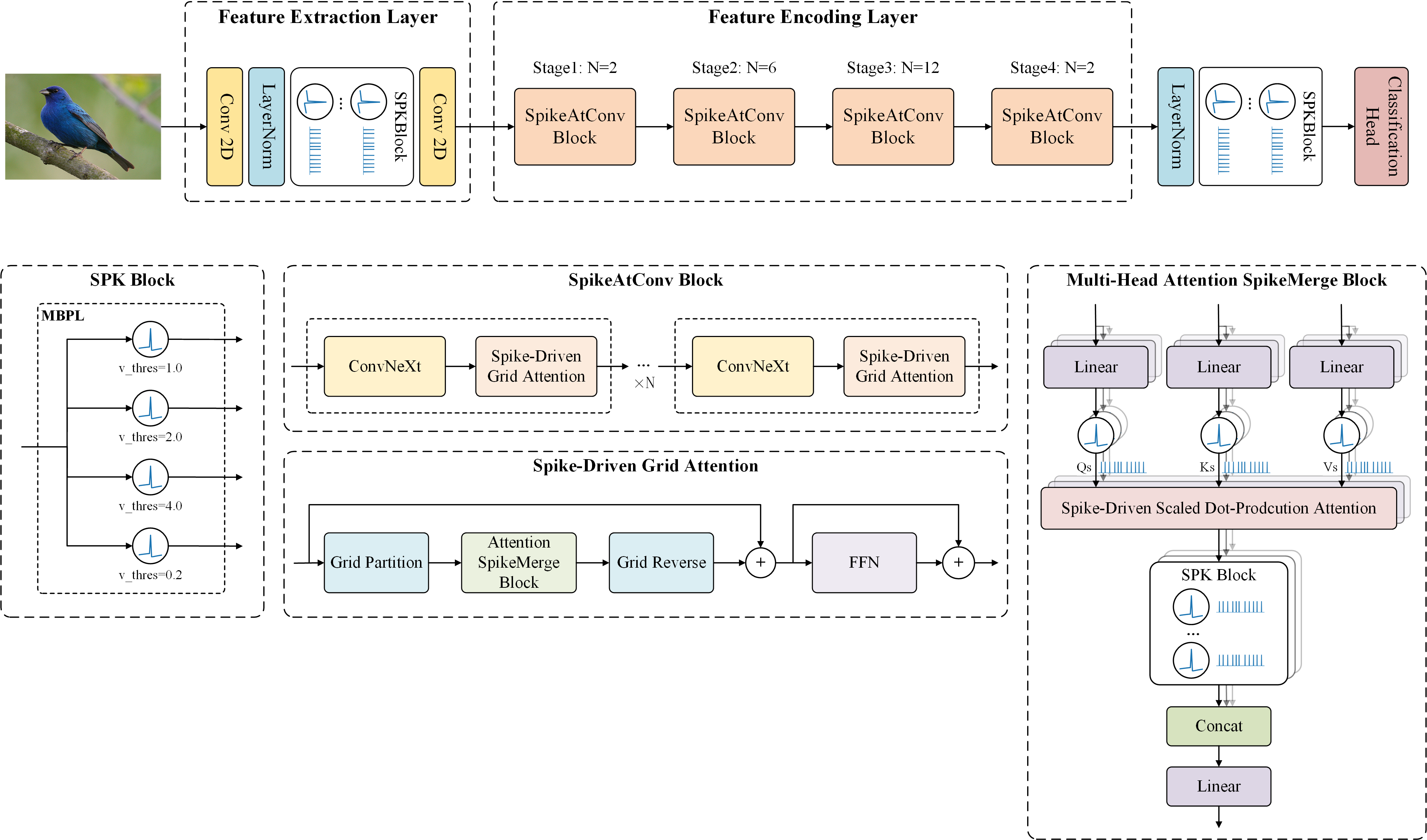}

  \caption{\textbf{The overview of SpikeAtConv.} The model is primarily composed of three components: the Feature Extraction Layer, the Feature Encoding Layer, and the Decision Layer. Initially, the input image is subjected to preliminary processing within the Feature Extraction Layer, where essential characteristics are identified. Subsequently, the Feature Encoding Layer performs a comprehensive analysis to distill salient features from the extracted data. Finally, the decision layer synthesizes this information to generate the prediction results.}
  \label{fig:SpikeAtConv}
\end{figure}

\subsection{SPK Block}

The Leaky Integrate-and-Fire (LIF) neuron model is a fundamental computational neuroscience model, widely used for its simplicity and reasonable approximation of biological neuron behavior. The core of the LIF model lies in simulating the dynamics of the neuronal membrane potential, which is governed by the following differential equation:

\begin{equation}
\tau_m \frac{dV}{dt} = -(V - V_{\text{rest}}) + RI(t)
\end{equation}

where $V$ represents the membrane potential, $\tau_m$ is the membrane time constant, $V_{rest}$ is the resting membrane potential, $R$ is the membrane resistance, and $I(t)$ is the input current. When the membrane potential $V$ exceeds the threshold $V_{threshold}$, the neuron fires a spike, and the membrane potential is reset to a lower reset potential 
$v_{reset}$, after which the neuron enters a refractory period during which it is unresponsive to new inputs. Other important hyperparameters in the LIF model include the duration of the refractory period, which affects the firing frequency and the response to consecutive inputs.

Leveraging the dynamic properties of LIF neurons, we have designed five SPK Blocks, as illustrated in Figure 2, to emulate various aspects of biological neural network information processing mechanisms.

\textbf{Single-Layer LIF SPK (SL) Block}: This fundamental building block consists of a single LIF neuron. Despite its simplicity, it effectively simulates the activation and inhibition dynamics of an individual neuron.

\textbf{Residual LIF SPK (RL) Block}: In this design, features pass through a LIF neuron and then split into two branches. The main branch is processed by a second LIF neuron, while the auxiliary branch retains the original features. The residual connection mitigates information loss and enhances the learning capacity of the model. 

\textbf{Multi-Branch Parallel LIF SPK (MBPL) Block}: This block comprises several LIF neurons with distinct hyperparameters arranged in parallel, allowing features to pass through several different thresholds simultaneously. The outputs of these neurons are then combined and fed into a ConvNeXt module to simulate membrane potential variations before summing the results. This approach enables the model to integrate information across different scales effectively.

\textbf{Hidden Split LIF SPK (HSL) Block}: This design splits and concatenates the outputs of two neurons along the hidden dimension. This method allows the model to capture features across different representational space dimensions, enhancing the model's expressive power.

\textbf{Dual Convolutional LIF SPK (DCL) Block}: The Dual Convolutional LIF SPK (DCL) Block features a bifurcated architecture with two parallel branches, each comprising a convolutional layer and a LIF neuron. The first branch harnesses a 3x3 convolutional kernel to discern fine spatial details, whereas the second branch leverages a 5x5 kernel to apprehend a wider spatial context. This strategy of extracting features at varying scales enables the DCL Block to simultaneously process spatial details with high and low resolution. Each branch's convolutional layer halves the channel dimension, and the outputs from both branches are subsequently concatenated along the channel axis, maintaining the original dimensionality.

\begin{figure}
  \centering
  \includegraphics[width=0.6\textwidth]{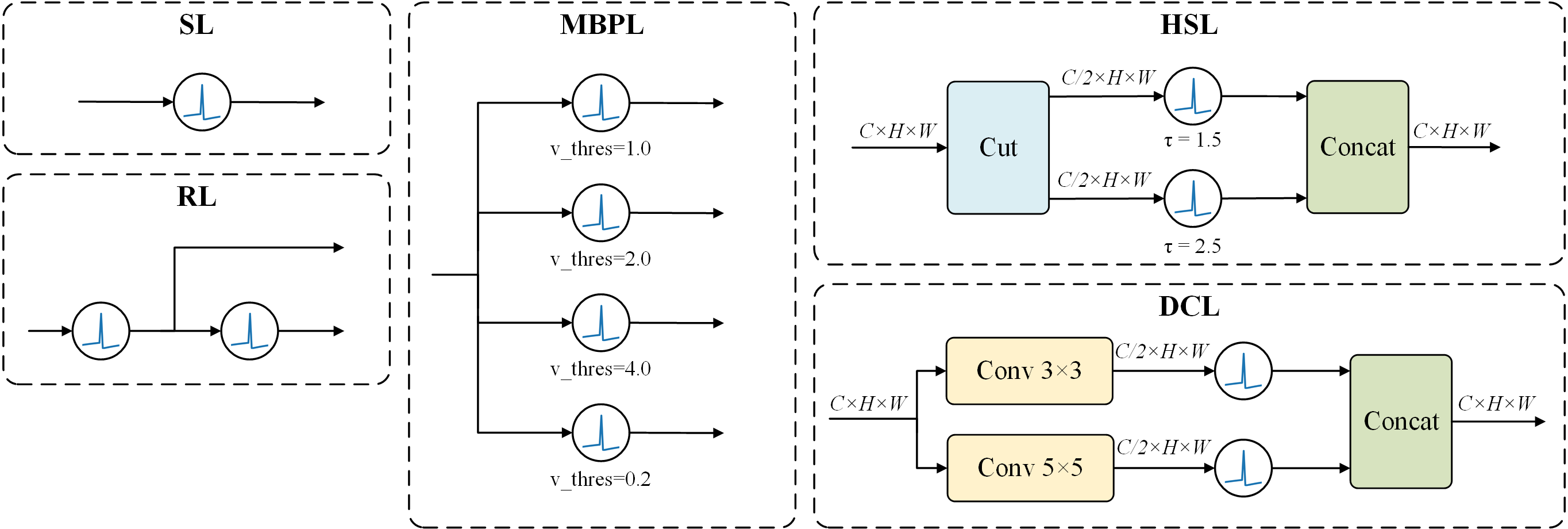}
  \caption{\textbf{SPKBlock.} Based on LIF neurons, we designed multiple SPK blocks to explore the impact of various hyperparameters and different combinations of multiple neurons on network performance. For example, the MBPL Block consists of multiple parallel neurons with different thresholds, while the DCL Block is composed of two parallel branches, each including a convolutional layer and a LIF neuron.}

  \label{fig:SPK Block}
\end{figure}

\subsection{Attention SpikeMerge Block}

ViT was a pioneering effort to apply a pure Transformer architecture to image recognition, demonstrating the impressive capabilities of Transformers in image processing. However, ViT also revealed several challenges, such as optimization difficulties, convergence issues, and high computational and memory costs. Additionally, handling long-tail effects, intra-class variations, and designing effective positional encodings remain areas requiring further investigation.

MaxViT addresses these issues by incorporating the multi-axis self-attention (Max-SA) module, which balances local and global attention. The Max-SA module combines window attention with grid attention, providing a better inductive bias, and uses CNNs for positional encoding, thereby mitigating some of ViT's limitations.

Building on the MaxViT architecture, we propose two distinct Attention SpikeMerge Blocks that integrate the Spk module to process spike signals. Our goal is to optimize the combination of attention mechanisms with Spk modules based on Leaky Integrate-and-Fire (LIF) neurons to enhance spike signal processing.

\textbf{Spike-Integrated Self-Attention (SISA) Block}: In this approach, the SPK Block is incorporated during the computation of the self-attention query (Q), key (K), and value (V). After calculating the attention scores and applying them to V, the SPK Block converts the attention map into spike signals. In the Feed-Forward Network, each linear layer is followed by an Spk module to maintain the spike-based processing.

\textbf{Binary Direct-Spike Attention (BDSA) Block}: This approach diverges significantly from the previous one. Given the binary nature of spike signals, we bypass the computation of Q, K, and V and directly transform the input features into spike signals, treating Q, K, and V as identical. This method accelerates the computation of the attention map by using matrix multiplication with identical binary vectors. During inference, post-training, the attention-processed features can be directly obtained without additional computation, simplifying the process.

We trained models using these two distinct Attention SpikeMerge Blocks to evaluate the impact of varying module complexity on model performance. The experimental results and their implications will be discussed in the following sections.


\begin{figure}
  \centering
  \includegraphics[width=0.8\textwidth]{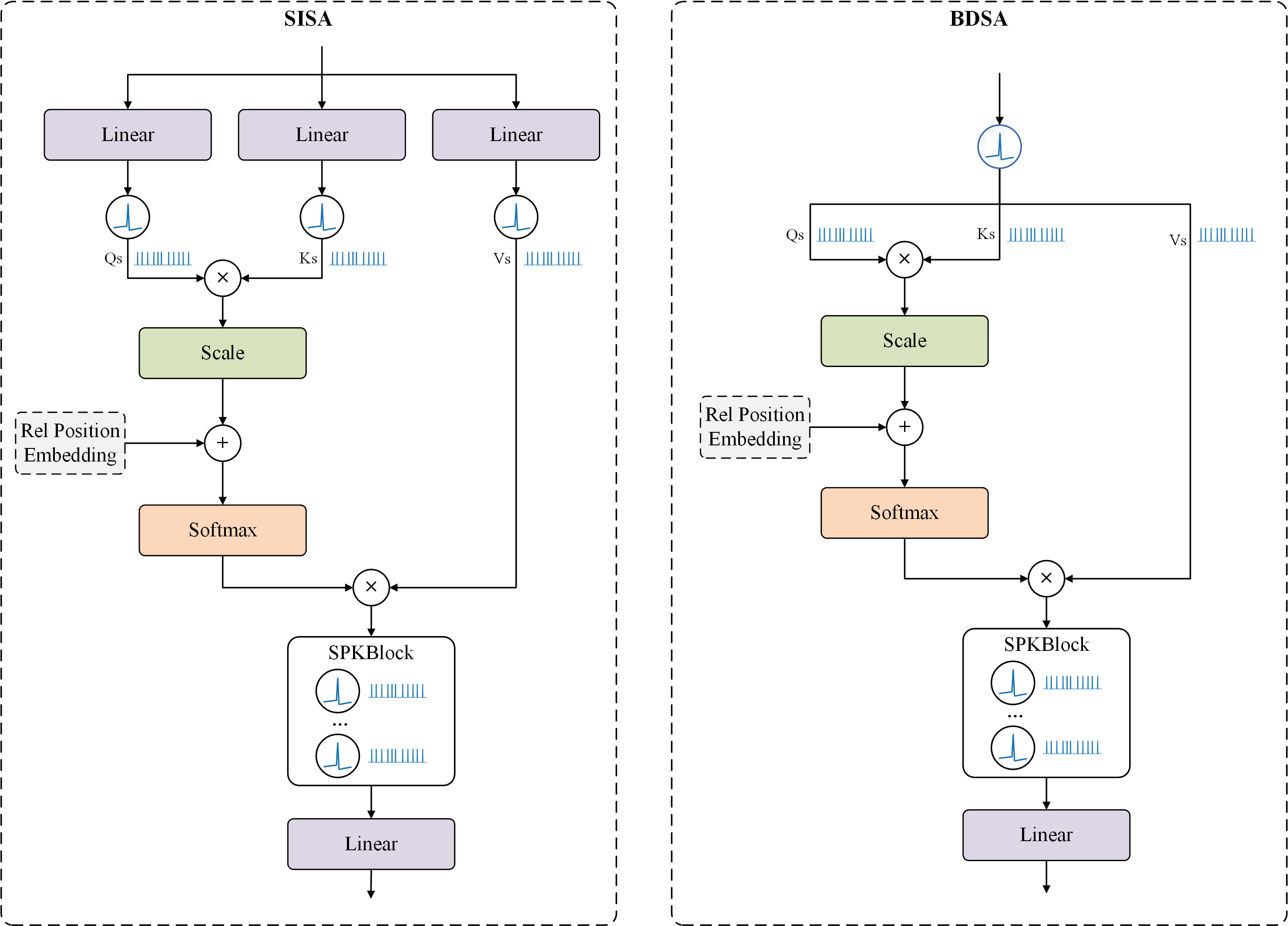}
  \caption{\textbf{Attention SpikeMerge Block.} This represents two different computational approaches. In the SISA Block, after computing the Q, K, and V, we add the SPK Blcok separately to obtain the spike form of Q, K, and V. Subsequently, we use Q and K to calculate the attention scores, apply these scores to V, and then incorporate the SPK Block to convert the attention into spike sequences. In the BDSA Block, we bypass the computation of Q, K, and V, directly converting the input  into spike sequences through the SPK Block, treating Q, K, and V as the same.}
  \label{fig:SPKAttention}
\end{figure}

\section{EXPERIMENTS}

\subsection{CIFAR-100 Experiments}

In this study, we utilized the CIFAR-100 dataset as a preliminary benchmark to evaluate and refine the design of various SPKBlock modules. CIFAR-100 is a well-regarded image classification dataset comprising 60,000 images across 100 distinct categories \cite{krizhevsky2009learning}. Its manageable scale and diversity make it an ideal choice for initial experimentation aimed at optimizing computational resources and reducing experimental time.

To highlight the impact of SPKBlock modules, we deliberately refrained from optimizing the training parameters, instead opting for a straightforward and commonly used set of settings. Specifically, training was conducted over 200 epochs, AdamW optimizer with an initial learning rate of 0.02\cite{loshchilov2017decoupled}. We opted for a batch size of 256 to ensure efficient use of computational resources while maintaining reasonable memory consumption. To mitigate early training instability, a warm-up strategy was implemented during the first 5 epochs. The weight decay parameter was set at 0.01 to counteract overfitting. Our primary goal was not to achieve a highly optimized model on CIFAR-100, but rather to evaluate the effectiveness of different SPKBlock modules.

In terms of data augmentation, we applied horizontal flipping, random rotations within a 30-degree range, and random shearing to enhance the model's generalization capabilities. Our base model architecture was ResNet-18, with various SPKBlock modules replacing the traditional activation functions to explore their impact. We systematically adjusted hyperparameters such as surrogate functions, voltage thresholds, tau values, and the spatial and temporal configurations of LIF neurons.

\subsection{ImageNet1K Experiments}

In this study, we utilized the ImageNet-1K dataset as a benchmark to evaluate and compare the efficacy of different neural network module designs \cite{deng2009imagenet}. ImageNet-1K is a widely-used image classification dataset that contains over one million annotated images across one thousand distinct categories. Its diversity and scale make it a significant challenge in the field of computer vision.

Regarding our experimental setup, we employed a series of meticulously chosen training parameters. Specifically, we set our training to run for 200 epochs to ensure ample learning opportunities. The initial learning rate was set at 0.001, a value aimed at balancing convergence speed and training stability. We opted for a batch size of 768 to make efficient use of our computational resources while maintaining reasonable memory consumption. During the first 10 epochs, we implemented a warm-up strategy to mitigate early training instability. The weight decay parameter was set at 0.05 to help counteract overfitting. The gradient clipping threshold was established at 0.1 to prevent gradient explosion issues. All images were resized to a uniform resolution of 224×224 to maintain consistency in input data.

Additionally, we adopted a cosine learning rate decay strategy, which allows for a smooth reduction of the learning rate in the later stages of training, aiding the model in converging to a more optimal solution. For data augmentation, we utilized the AutoAugment technique, an approach that optimizes augmentation policies through automatic searching. We also employed label smoothing with a value of 0.1 to reduce the model's sensitivity to label noise. Techniques such as Random Erase, Mixup, and CutMix were integrated as well, which have been proven to effectively enhance the model's generalization capabilities on images \cite{zhang2020does,yun2019cutmix}.

We designed three different model architectures to explore the impact of varying network scales on performance: Tiny, Base, and Large models.

For the \textbf{Tiny} model, the hidden state dimensions were set to 64, 128, 256, and 512 for the four stages, respectively. The module depths for each stage were set at 1, 3, 6, and 1. This configuration aims to provide a lightweight model suitable for environments with limited computational resources.

For the \textbf{Base} model, the hidden state dimensions were set to 128, 256, 512, and 1024 for the four stages, respectively. The module depths for each stage were set at 2, 6, 12, and 2. This design is intended to progressively extract and process features of the images, balancing computational efficiency and performance.

For the \textbf{Large} model, the hidden state dimensions were set to 160, 320, 640, and 1280 for the four stages, respectively. The module depths for each stage were set at 2, 6, 16, and 2. This configuration aims to capture more complex features and provide higher accuracy, suitable for environments where computational resources are abundant.

\subsection{Main Properties}

We ablate SpikeAtConv using the default settings from section 4.1 and observe some interesting phenomena.

\subsubsection*{Comparative Analysis of SPK Block Variants}

To evaluate the performance of different SPK blocks, we selected SISA as the Attention SpikeMerge Block. Our experiments on the CIFAR-100 dataset systematically assessed various SPKBlock configurations, focusing on the impact of different simulation time windows, surrogate gradient functions, and the number of branches on model accuracy. Table \ref{table:cifar100_spkblock_acc} presents a detailed comparison of these configurations.

The analysis of Single-Layer (SL) configurations revealed that increasing the simulation time window \( T \) generally enhances model performance. For instance, the accuracy improved from 64.3\% at \( T = 1 \) to 70.9\% at \( T = 4 \). However, when \( T \) was further increased to \( T = 8 \), there was a slight performance drop to 70.4\%. This indicates an optimal range for \( T \), beyond which the benefits diminish, likely due to increased complexity and potential overfitting.

Additionally, the choice of surrogate gradient functions significantly impacted performance. The Atan function consistently outperformed the Sigmoid function under equivalent settings. For example, with \( T = 1 \) and \(\tau = 2.0\), the accuracy with Atan was 67.1\%, compared to 66.0\% with Sigmoid. This suggests that the Atan function provides a more effective gradient approximation for training spiking neurons.

In the MBPL Block experiments (see Table \ref{table:MBPL_acc}), we observed that increasing the number of branches markedly enhanced performance. A configuration with four branches and varied voltage thresholds achieved a top-1 accuracy of 74.4\%, significantly higher than simpler configurations. However, an excessive number of branches, such as eight, resulted in a performance drop to 59.0\%. This decline was attributed to the insufficient training of the numerous branches, which introduced noise and hampered the model's learning capacity.

In our CIFAR-100 experiments, we only listed representative examples. For the SSL and HSL modules, when using the same number of LIF branches, their performance was similar to that of the MBPL. However, they are either computationally more complex or less scalable, so we did not list more detailed results.

\begin{table}[!htbp]
  \caption{\textbf{Accuracy for different SPKBlock on CIFAR-100.} Surrogate denotes surrogate gradient, a smooth approximation used to train spiking neural networks.}
 \centering
 \begin{tabular}{llllllll}
    \toprule
    Model  & Top-1 Acc (\%) & SPKBlock  & $T$ & $\tau$ & V Threshold & Number of Branches & Surrogate \\
    \midrule
    ResNet-18 & 75.4 & --- & --- & --- & --- & --- & ---  \\
    ResNet-18 & 64.3 & SL & 1 & 4.0 & 1.0 & 1 & Sigmoid  \\
    ResNet-18 & 66.0 & SL & 1 & 2.0 & 1.0 & 1 & Sigmoid  \\
    ResNet-18 & 67.1 & SL & 1 & 2.0 & 1.0 & 1 & ATan  \\
    ResNet-18 & 70.9 & SL & 4 & 2.0 & 1.0 & 1 & Sigmoid  \\
    ResNet-18 & 70.4 & SL & 8 & 2.0 & 1.0 & 1 & Sigmoid  \\
    ResNet-18 & 71.0 & RL & 1 & 2.0 & 1,2 & 2 & ATan  \\
    ResNet-18 & 74.4 & MBPL & 2 & 2.0 & 0.2,1,2,4 & 4 & ATan  \\
    ResNet-18 & 70.6 & HSL & 2 & 2.0 & 1,2 & 2 & ATan  \\
    ResNet-18 & 70.3 & DCL & 2 & 2.0 & 1,1 & 2 & ATan  \\
    \bottomrule
  \end{tabular}
 \label{table:cifar100_spkblock_acc}
\end{table}

\begin{table}[!htbp]
\caption{\textbf{MBPL Block experiments on CIFAR-100.}}
 \centering
 \begin{tabular}{llllllll}
    \toprule
    Model  & Top-1 Acc (\%) & SPKBlock  & $T$ & $\tau$ & V Threshold & Number of Branches & Surrogate \\
    \midrule
    ResNet-18 & 71.7 & MBPL & 1 & 2.0 & 1,2 & 2 & ATan  \\
    ResNet-18 & 71.9 & MBPL & 1 & 2.0 & 1,2,4 & 3 & ATan  \\
    ResNet-18 & 72.5 & MBPL & 1 & 2.0 & 0.2,1,2,4 & 4 & ATan  \\
    ResNet-18 & 72.2 & MBPL & 1 & 2.0 & 1,2,4,6 & 4 & ATan  \\
    ResNet-18 & 73.8 & MBPL & 2 & 2.0 & 0.2,1,2,4 & 4 & Sigmoid  \\
    ResNet-18 & 74.4 & MBPL & 2 & 2.0 & 0.2,1,2,4 & 4 & ATan  \\
    ResNet-18 & 73.6 & MBPL & 4 & 2.0 & 0.2,1,2,4 & 4 & ATan  \\
    ResNet-18 & 59.0 & MBPL & 2 & 2.0 & 0.2,1,2,3,4,5,6,7 & 8 & ATan  \\
    \bottomrule
  \end{tabular}
 \label{table:MBPL_acc}
\end{table}

Through extensive experimentation, we discovered that the MBPL module exhibited the best overall performance. For each module, we identified the optimal hyperparameter settings and structural configurations specific to CIFAR-100. These preliminary results allowed us to eliminate numerous suboptimal designs and provided valuable insights for further experiments on more complex datasets like ImageNet-1K.

Table \ref{table:spkblock_acc} presents a performance comparison of various SPKBlock configurations on the ImageNet-1K dataset. It is observed that the MBPL modules outperform others, while the SL module exhibits comparatively weaker performance. This phenomenon suggests a disparity in the information capture capabilities of different LIF neurons: a combination of multiple LIF neurons appears to integrate a richer set of information, compensating for the potential deficiencies of individual LIF neurons in processing information.

Additionally, it is noted that when information sequentially passes through two LIF neurons, there seems to be a reduction in the amount of effective information in the output. This could explain why the RL module performs better than the SL module but still falls short of the SSL. 

However, despite the theoretical advantages, we found that DCL modules prevented the model from being fully trained. After 40 epochs, the loss of the training set stopped decreasing. Even lowering the learning rate, adjusting the position of the SPK module, or adding a normalization layer did not resolve this issue. This is a common problem encountered when processing spiking signals, where the neural spike module is prone to crashing in the existing deep learning training framework.

\begin{table}[!htbp]
  \caption{\textbf{Accuracy for different SPKBlock on ImageNet1K.}}
 \centering
 \begin{tabular}{lllllll}
    \toprule
      & SL  & RL & MBPL & HSL & DCL & spike-free \\
    \midrule
    Top1 (\%) & 74.91 & 78.35 & 80.53 & 78.44 & 77.66 & 81.13 \\
    Top5 (\%) & 91.94 & 93.73 & 94.17 & 93.75 & 93.83 & 94.30 \\
    \bottomrule
  \end{tabular}
 \label{table:spkblock_acc}
\end{table}

\subsubsection*{Comparison of SpikeAtConv and Other Models on ImageNet-1K}

We evaluated the performance of our SpikeAtConv model at different scales (Tiny, Base, and Large) on the ImageNet-1K dataset. Each model was trained for 200 epochs with an image input resolution of 224. Most LIF neurons were set with a time dimension of 1; however, we added an experiment with the Base model using T=2 to explore its impact.  Our findings indicate that while the Large model achieves slightly higher accuracy than the Base model, further improvements are expected with increased image resolution and additional training epochs.

Table \ref{table:scale_acc} presents a detailed comparison of the performance metrics, including top-1 and top-5 accuracy, as well as the number of parameters for each model. The SpikeAtConv models are also compared against state-of-the-art models like Meta-SpikeFormer and SpikFormer.

\begin{table}[!htbp]
  \caption{\textbf{Performance of different models on ImageNet-1K.}}
  \centering
  \begin{tabular}{lllllll}
    \toprule
      & Base(T=1)  & Base(T=2) & Tiny(T=1) & Large(T=1) & Meta(T=1)  & SpikFormer(T=1) \\
    \midrule
    Top-1 (\%) & 80.53 & 80.70 & 76.58 & 81.23 & 79.1 & 74.8 \\
    Top-5 (\%) & 94.17 & 94.89 & 92.74 & 95.41 & ---  &  --- \\
    \bottomrule
  \end{tabular}
  \label{table:scale_acc}
\end{table}

From the table, it is evident that our Large SpikeAtConv model achieves a top-1 accuracy of 81.23\%, outperforming both Meta-SpikeFormer (79.1\%) and SpikFormer (74.8\%). The Base model also shows competitive performance with a top-1 accuracy of 80.53\%. Despite having fewer parameters, the Tiny model maintains a respectable top-1 accuracy of 76.58\%, demonstrating the efficiency of our approach.In terms of top-5 accuracy, the Large SpikeAtConv model reaches 95.41\%, and the Base model achieves 94.17\%.

The comparison indicates that SpikeAtConv models, particularly the Large variant, provide superior performance on ImageNet-1K, while maintaining a balance between accuracy and model complexity. This demonstrates the effectiveness of our approach in leveraging spiking neural networks for large-scale image classification tasks.

\section*{Discussion}

In summary, our results highlight two key points. First, a well-designed SNN architecture can significantly enhance the performance of spiking neural networks (SNNs). Second, integrating SNNs with advanced deep learning architectures can further improve their performance.

For the first point, we observed that SNN modules based on LIF neurons tend to lose a considerable amount of information. While redundant picture information might be unnecessary for classification tasks, the SNN module acts similarly to an activation function, filtering out irrelevant details. However, this filtering can also lead to performance degradation. To address this, we adopted an approach akin to early convolutional neural networks (CNNs) by setting up parallel LIF neurons with different parameters. This setup captures different levels of information, thereby maximizing the richness of information extracted by the SNN.

\begin{figure}
  \centering
  \includegraphics[width=0.8\textwidth]{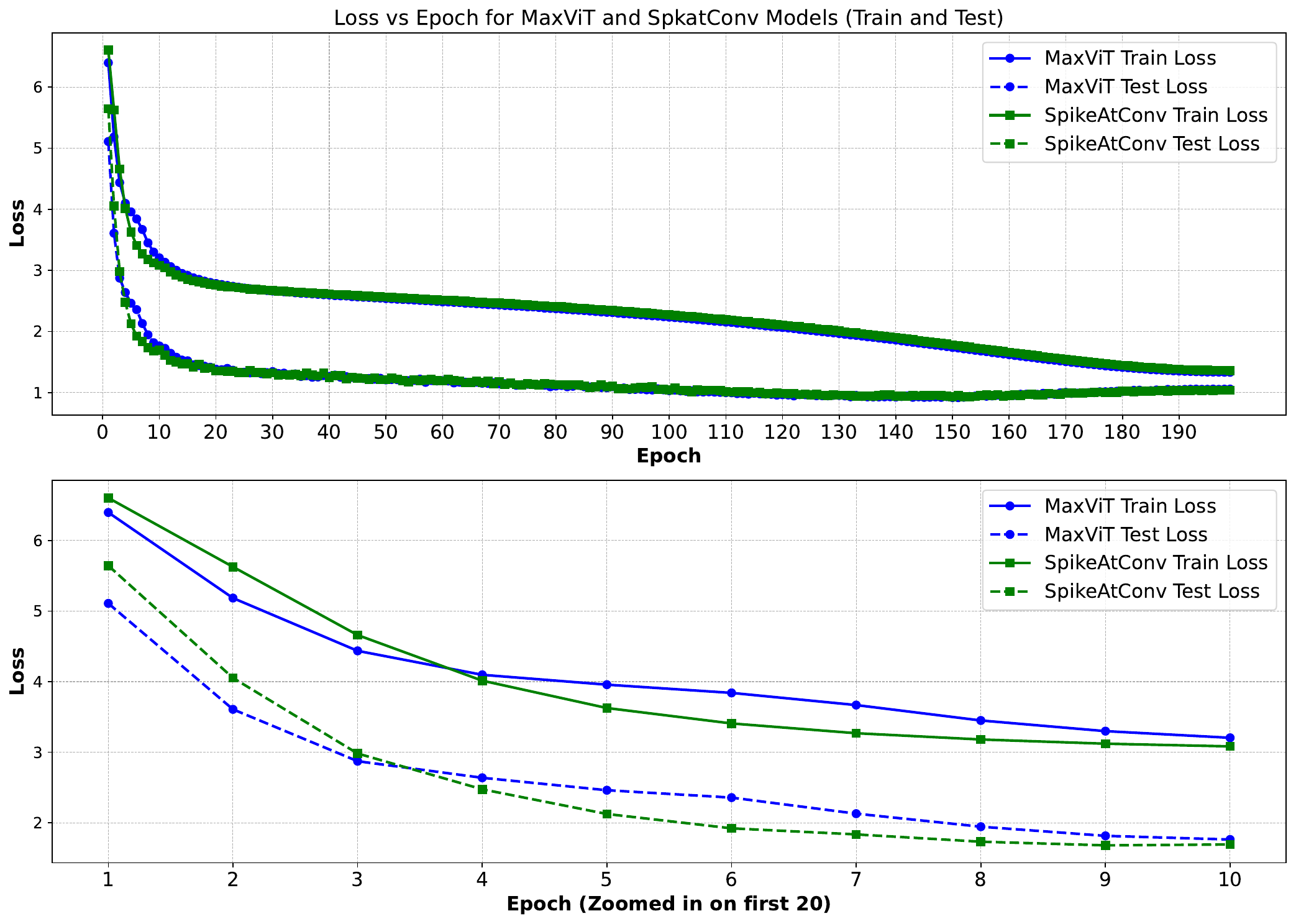}
  \caption{\textbf{Comparison of Loss Between MaxViT  and SpikeAtConv.} We present the training and validation loss trajectories of our SpikeAtConv and MaxViT models. In the figure below, we emphasize the loss variations during the first 10 epochs. It is evident that SpikeAtConv experiences a slower reduction in loss during the initial 5 epochs. Due to the application of data augmentation techniques such as auto augment and mixup during training, the training loss consistently remains higher than the test loss.}
  \label{fig:loss_compare}
\end{figure}

Regarding the second point, our experiments demonstrate that the reasonable integration of SNN modules can have minimal impact on the original performance of the neural network. However, we also noticed that the loss of the SNN network decreased more slowly in the early stages of training compared to networks without SNNs (Figure \ref{fig:loss_compare}). This suggests that while SNNs have the potential to achieve excellent results in visual tasks, further research is needed to develop corresponding training methods and network modules that effectively cooperate with SNNs.

Traditional deep learning architectures like CNNs and Transformers have been extensively researched, leading to the development of numerous auxiliary layers, targeted data augmentation techniques, and pre-training strategies that ensure these networks are well-trained and stable. Similarly, SNNs require further in-depth studies to develop analogous methods that can ensure sufficient training and stability during the training process. This includes designing specialized layers, data augmentation techniques, and training protocols tailored specifically for SNNs to unlock their full potential.

\section{Conclusion}

In this study, we developed a novel spiking neural network (SNN) model named SpikeAtConv, which achieved state-of-the-art (SOTA) results among SNN models on the ImageNet-1K dataset. Our approach involved designing a series of Spk blocks to convert continuous hidden states into neural spikes. Through extensive experimentation, we identified the optimal Spk block configuration and integrated it with the MaxVit architecture. This combination enabled us to significantly advance the performance of SNNs.

One of our key findings was that even when using a degenerate self-attention mechanism, the performance of our model did not degrade significantly. This suggests that our Spk blocks are highly effective in capturing and processing information, even without the full complexity of self-attention.

Additionally, our experiments demonstrated that a well-designed SNN architecture can substantially enhance performance. By setting up parallel LIF neurons with different parameters, we were able to capture various levels of information, thereby enriching the data representation within the SNN.

Looking forward, we aim to further refine the design of Spk blocks and explore improvements in backpropagation techniques. These enhancements will help ensure that Spk blocks are fully trained and can further improve the performance and stability of SNNs across various tasks. Moreover, we recognize the need for developing specialized training methods, auxiliary layers, and data augmentation techniques tailored specifically for SNNs, akin to the extensive research conducted for CNNs and Transformers.

In conclusion, our work not only introduces a powerful new SNN model but also lays the groundwork for future research in optimizing SNN architectures and training methodologies. We believe that with continued exploration and innovation, SNNs can achieve even greater performance and applicability in diverse domains.

\bibliographystyle{unsrt}  
\bibliography{references}

\end{document}